\documentclass[letterpaper, 10 pt, conference]{ieeeconf}  % Comment this line out if you need a4paper

\IEEEoverridecommandlockouts                              % This command is only needed if 
\title{\LARGE \bf
AI Olympics challenge with Evolutionary Soft Actor Critic
}

\author{Marco Calì$^{1}$, Alberto Sinigaglia$^{2}$, Niccolò Turcato$^{1}$, Ruggero Carli$^{1}$ and Gian Antonio Susto$^{1,}$$^{2}$% <-this % stops a space
\thanks{*This work was not supported by any organization}% <-this % stops a space
\thanks{$^{1}$Department of Information Engineering, University of Padova, Italy.
        {\tt\small marco.cali.2@studenti.unipd.it}}%
\thanks{$^{2}$Human-Inspired Technology Research Center, University of Padova, Via Luzzatti, 4, Padova, 35121, Italy
        {\tt\small alberto.sinigaglia@phd.unipd.it}}%
}

\usepackage{xcolor}
\usepackage{amsmath}
\usepackage{amsfonts}
\usepackage{graphicx} 
\usepackage{url} 
\usepackage{cleveref}

\begin{document}

\maketitle
\thispagestyle{empty}
\pagestyle{empty}

%%%%%%%%%%%%%%%%%%%%%%%%%%%%%%%%%%%%%%%%%%%%%%%%%%%%%%%%%%%%%%%%%%%%%%%%%%%%%%%%
\begin{abstract}

In the following report, we describe the solution we propose for the AI Olympics competition held at IROS 2024. Our solution is based on a Model-free Deep Reinforcement Learning approach combined with an evolutionary strategy. We will briefly describe the algorithms that have been used and then provide details of the approach.

\end{abstract}

%%%%%%%%%%%%%%%%%%%%%%%%%%%%%%%%%%%%%%%%%%%%%%%%%%%%%%%%%%%%%%%%%%%%%%%%%%%%%%%%
\section{Introduction}
In this paper, we present the approach taken for solving the simulation stage of the AI Olympics competition held in IROS 2024. Our approach first uses Model-Free Reinforcement Learning to find a policy that performs the swing up and stabilization of the robots, and then finetunes the agents through evolutionary methods. In particular, it first trains the agent using the Soft Actor-Critic (SAC)\cite{sac} algorithm with a physics-inspired reward function in order to train an agent that is able to perform the main task. Thanks to the ability of SAC to have stable yet effective training of the agents, we are able to optimize a reward function that approximates the competition's reward function with just a few hours of training while obtaining results that surpass the proposed baselines. Furthermore, SAC trains the agent to be robust out of the box, thanks to its algorithmic specifications.

Finally, the agent undergoes further optimization through the application of evolutionary algorithms to optimize the competition's desired score function.

This paper is organized as follows: Section 2 introduces the challenge, Section 3 introduces the Soft Actor-Critic algorithm, Section 4 instead introduces evolutionary strategies, Sections 5, 6, and 7 introduce the reward and training schedule used to train the final agent, Section 8 introduces the experiments settings used for the simulation training, while Section 9 highlights the challenges of real robot experiments. Finally, Section 10 concludes the paper.

The submission code can be found at \url{https://github.com/AlbertoSinigaglia/double_pendulum}

% \begin{figure}
%     \centering	
%     \includegraphics[width=0.35\columnwidth]{images/Acrobot.png}%
%     \includegraphics[width=0.35\columnwidth]{images/schematic.png}
%     \caption{The double pendulum, system from RealAIGym: real hardware, schematic.}
%     \label{fig:system}
% \end{figure}

\section{Competition's goal}

The competition involves an underactuated double pendulum system with 2 degrees of freedom\cite{doublependolumsystem}, combined in two settings named \textit{acrobot} and \textit{pendubot}, whose state is represented as $(\theta_1, \theta_2, \dot \theta_1, \dot \theta_2)$. In the former, the first joint is passive while the second one is actuated, whereas in the latter, the first joint is actuated, and the second one is passive.

The competition is composed of a simulation and a real hardware stage. 

In the simulation stage, the goal is to derive a controller for each setting that performs the swing-up and stabilization in the vertical setting, an unstable equilibrium point of the systems. Both systems are simulated with a frequency of $500 \, \text{Hz}$ for $10$ seconds.
Finally, the proposed controllers are also tested for their robustness. For this stage, an OpenAI Gym environment is employed \cite{openaigym}. Previous years' competition submissions can be found in \cite{oldcompetition}.

In the real hardware stage, controllers are tested on the physical systems.

In the rest of the sections, we review the simulation stage's performance and robustness evaluations.

\section{Soft Actor Critic}
This section provides a concise review of the Soft Actor-Critic (SAC) algorithm, its origins, and its training routine.

\subsection{Deep Deterministic Policy Gradient (DDPG)}
Expanding on the concepts of Deterministic Policy Gradient (DPG) \cite{dpg}, the Deep Deterministic Policy Gradient (DDPG) algorithm \cite{ddpg} integrates deep learning. DDPG employs two deep neural networks: the actor and the critic. The actor-network directly maps states to actions, while the critic network approximates the Q-function. The core update equations in DDPG are as follows:
Critic update: 
\begin{align}
    &\Delta \phi = \mathbb{E}_{(s, a, r, s')} \left[ (r + \gamma Q_{\phi'}(s', \pi_{\theta'}(s')) - Q_\phi(s, a))^2 \right]
\end{align}
Actor update:
\begin{align}
    &\Delta \theta = \mathbb{E}_{s} \left[ \nabla_{a} Q_\phi(s, a) |_{a=\pi_\theta(s)} \nabla_{\theta} \pi_\theta(s) \right]
\end{align}
where \(\phi\) and \(\theta\) are the parameters of the critic and actor networks, respectively. DDPG also uses a replay buffer and target networks (\(\phi'\) and \(\theta'\)), enhancing stability and efficiency.

\subsection{Soft Actor-Critic (SAC)}
The Soft Actor-Critic (SAC) algorithm, presented in \cite{sac}, introduces an entropy term to the reward structure, promoting exploration and robustness in policy learning. SAC optimizes a stochastic policy in a way that balances the trade-off between entropy and reward. The objective function for SAC is given by:
\begin{equation}
    J(\pi) = \mathbb{E}_{s_t, a_t \sim \pi} \left[ \sum_{t} \gamma^t \left( r(s_t, a_t) + \alpha \mathcal{H}(\pi(\cdot | s_t)) \right) \right],
\end{equation}
where \(\mathcal{H}\) denotes the entropy of the policy \(\pi\) and \(\alpha\) is a temperature parameter that controls the importance of the entropy term against the reward. SAC employs separate networks for the policy, value function, and two Q-functions, reducing overestimation bias and improving learning stability.

\section{Zero-Order Optimization and Evolutionary Strategies}
Zero-order optimization, or derivative-free optimization, encompasses a range of techniques that optimize functions without requiring gradient information. These methods are essential in scenarios where gradients are non-existent, expensive to compute, or do not reliably guide the optimization due to noise, discontinuities, or non-differentiability. Evolutionary strategies (ES) are a prominent class of zero-order optimization techniques that employ mechanisms analogous to biological evolution, such as mutation, recombination, and selection, to evolve a population of candidate solutions over successive generations.

% \subsection{Comparison with Finite Difference Methods}
% Finite difference methods approximate gradients by evaluating the function at two or more closely spaced points and calculating the difference quotient. Despite their simplicity, these methods can be inefficient and inaccurate, particularly in high-dimensional spaces or when the objective function evaluations are noisy or costly. Evolutionary strategies offer an alternative by using only function values (fitness scores) to guide the search process, eliminating the need for explicit gradient estimation. This characteristic allows ES to be robust against the issues that complicate gradient-based methods, as illustrated by the following generic update formula often used in finite difference methods:
% \begin{equation}
%     \nabla f(\theta) \approx \frac{f(\theta + h) - f(\theta)}{h},
% \end{equation}
% where \( h \) is a small step size. In contrast, ES typically employs a population-based stochastic search that does not directly compute this gradient but instead uses population statistics to inform direction and magnitude of parameter updates.
    
\subsection{Separable Natural Evolution Strategy (SNES)}
The Separable Natural Evolution Strategy (SNES) \cite{snes} refines evolutionary strategies by focusing on the efficient adaptation of mutation distributions. SNES is particularly effective in environments where parameters have different scales and sensitivities. It maintains and adapts a separate step size for each parameter dimension, facilitating an independent and efficient exploration of the parameter space. The core of the SNES algorithm involves updating the mutation strengths using a log-normal rule, which is mathematically depicted as follows:
\begin{align}
    \sigma_{\text{new},i} &= \sigma_{\text{old},i} \exp \left( \tau \mathcal{N}(0, 1) + \tau' \mathcal{N}(0, 1)_i \right), \\
    \theta_{\text{new},i} &= \theta_{\text{old},i} + \sigma_{\text{new},i} \mathcal{N}(0, 1)_i,
\end{align}
for each dimension \(i\), where \(\tau\) and \(\tau'\) are learning rates designed to control the global and individual adaptation speeds, respectively. This adaptation mechanism is inspired by the principles of natural evolution and covariance matrix adaptation, but it simplifies the adaptation process by treating the search space dimensions as separate entities. The SNES thus combines the robustness of evolutionary algorithms with the efficiency of adaptive step-size mechanisms, leading to faster convergence and reduced computational complexity compared to traditional ES and other sophisticated adaptation strategies like CMA-ES.

\section{Surrogate reward function}
The competition's reward function combines different aspects of the learned controllers, such as the time required for the swing up, the energy consumed, the torque required, and many others. It then combines them with different weights since they all have different scales. However, such a reward is challenging to maximize for a Deep RL agent that observes only the current immediate state of the system since it includes terms that depend on the overall trajectory, thus violating the RL definition of reward function $\mathcal{R}(s, a, s')$, if not using the whole trajectory as the state. For this reason, we developed a surrogate reward function that should act as a feasible counterpart but still be proximal to the competition score function.

The reward function being used is the following:

\begin{align}
R(s,a) = \begin{cases}
V + \alpha [1+ \cos(\theta_2)]^2 - \beta T & \text{ if } y > y_{th}\\
\,\,\,\,\,\,\,\,\,\,\,\,\,\,\,\,\,\,\,\,\,\,\,\,\,\,\,\,\,\,-\rho_1 a^2 - \phi_1 \Delta a\\
V - \rho_2 a^2 - \phi_2 \Delta a -\eta ||\dot{s}||^2  & \text{ otherwise }
\end{cases}
\end{align}

where $\alpha, \beta, \phi, \eta, \rho$ are just hyperparameters that are used to control the tradeoffs. $y_{th}$ is set to  $0.375 \,\text{m}$ for acrobot and $0.35 \, \text{m}$ for pendubot. In this formulation, $a$ is the normalized action, thus $a \in [-1, 1]$. $V$ is the potential energy of the system, and $T$ is the kinetic energy. $\Delta a $ is the difference between the current action and the previous action. $\dot s$ is the square norm of the angular velocities of the robot, thus $||\dot s||^2 = \dot \theta_1^2+ \dot\theta_2^2$. Finally, $\theta_2$ is the angle between the two robot links.

\section{On the agent robustness}
\subsection{SAC agent robustness}
Soft Actor-Critic is a very popular algorithm used to learn policy for continuous control. Its optimization objective is the following:
\begin{equation}
    J(\theta) = \sum_T \mathbb{E}_{(s,a)\sim \pi_\theta}[r(s,a) + \alpha \mathcal{H}(\pi_\theta(\cdot | s))]
\end{equation}
For this reason, SAC models $\pi_\theta(s) \sim N(\mu(s), \text{diag}[\Sigma(s)])$. Thanks to this definition, the model is already optimized to find a reasonably robust solution. Indeed, the agent has to integrate all actions that it can take in the next state, which are normally distributed, and it's optimized to reward high entropy distributions. For this reason, the solution found by SAC, will already be robust to noise on the action sampled.

\subsection{SNES agent robustness}
The SNES algorithm is used for this project to align the agent with the actual reward function rather than the surrogate one.

However, training the agent to optimize the reward function greedily has a major flaw: if earlier the SAC algorithm was making the agent robust out of the box, for SNES, nothing prevents the agent from cherrypicking a trajectory extremely effectively, but at the same time extremely brittle.

For this reason, the training is slightly modified in the following way:
\begin{enumerate}
 \item sample the greedy action from the SAC agent: $a := \pi_\theta(s)$
 \item undo the $\text{tanh}$ transformation $a := \text{atanh}(a)$
 \item introduce noise: $a := a + \epsilon, \,\, \epsilon \sim N(0, \sigma^2)$
 \item apply squashing: $a := \text{tanh}(a)$
\end{enumerate}
Indeed, it is not enough to only take a non-greedy action and sample from the posterior learned by the agent because nothing prevents such posterior from collapsing to a Dirac delta distribution. Instead, we measured the average $\sigma$ computed by SAC after training during a trajectory, which was about $0.01$, and we introduced it manually on the action.

\section{Agent training}
The following section describes our two-step training approach for the Evolutionary SAC controller.
\subsection{Main agent training}
The training of the Soft Actor-Critic (SAC) agent begins by using the previously defined reward function. One of the most impactful hyperparameters is the maximum torque $\tau_{max}$, which directly influences the agent's actions. In our experiments, we explored various maximum torque limits to balance learning effectiveness and energy efficiency.

The torque values tested included $1.5 \, \text{Nm}$, $3.0 \, \text{Nm}$, and $5.0 \, \text{Nm}$. After conducting extensive experiments, we established that the torque limit had a notable impact on both the training dynamics and the final performance of the agent.

With a torque limit of $1.5 , \text{Nm}$, we noticed that restricting the torque encouraged the agent to develop more energy-efficient solutions. By imposing tighter constraints on the force the agent could exert, the agent focused on optimizing its actions to achieve the task with minimal energy usage. However, this conservative approach also introduced significant challenges. The reduced torque limited the agent's ability to apply sufficient force when necessary, causing it to stabilize about equilibrium points such as $(\theta_1, \theta_2) = (\pi, \pi)$ or $(\theta_1, \theta_2) = (0, \pi)$.

Conversely, when the torque limit was increased to $5.0 \, \text{Nm}$, the agent was empowered to exert greater force, which initially appeared to accelerate the learning process. The agent could more easily overcome the local maxima of the reward function and execute the required actions with brute force, leading to quicker early-stage learning. However, this initial advantage proved to be deceptive. As training progressed, the agents that relied on this higher torque tended to overuse the available force, resulting in less efficient and less precise solutions. These agents displayed excessive and unnecessary movements, which made the agent unable to stabilize around the desired equilibrium after the swing-up task. 

The $3.0 \, \text{Nm}$ torque limit struck a balance between the extremes. This setting provides the agent with sufficient power to perform necessary actions effectively while maintaining a reasonable degree of energy efficiency. Agents trained with this torque limit managed to learn the task within a reasonable timeframe and achieved competitive performance without resorting to excessive energy consumption. The $3.0 , \text{Nm}$ torque limit allowed for effective learning without the drawbacks associated with the lower or higher torque extremes.

This formulation led us to solutions with a score of $0.504$ for the acrobot and $0.567$ for the pendubot.

The complete list of hyperparameters being used can be found in \cref{original_parameters}

\begin{table}[h]
\caption{Hyper-parameters used for the SAC training}
\label{original_parameters}
\begin{center}
\begin{tabular}{|c||c|c|}
\hline
Param. & Acrobot & Pendubot\\
\hline
 $y_{th}$ & 0.375 & 0.35 \\
 $\tau_{max}$ & 3 & 3 \\
 $\alpha$ & 2 & 2\\
 $\beta$ & 1 & 1\\
 $\rho_1 $ & 0.1 & 0.1 \\
 $\rho_2 $ & 0.02 & 0.02 \\
 $\phi_1$ & 0.15 & 0.15 \\
 $\phi_2$ & 0.15 & 0.15 \\
 $\eta$ & 0.02 & 0.02 \\
 lr & 0.001 & 0.001 \\
\hline
\end{tabular}
\end{center}
\end{table}

% \subsection{Agent finetuning}
% Since tuning the hyperparameters a priori was very challenging, we chose to first train the agent with loose constraints and then reinforce the ones that correlate with the properties of the solution that we ideally want to smooth. 

% Furthermore, in order to avoid catastrophic forgetting, we lowered the stepsize by one order of magnitude in order to only slowly move away from the solution found by the first stage of the training.

% \begin{table}[h]
% \caption{Hyper-parameters used for the finetuning}
% \label{finetuning_parameters}
% \begin{center}
% \begin{tabular}{|c||c|c|}
% \hline
% Param. & Acrobot & Pendubot\\
% \hline
%  $y_{th}$ & 0.375 & 0.35 \\
%  $\alpha$ & 6 & 3\\
%  $\beta$ & 1 & 1\\
%  $\rho_1 $ & 0.3 & 0.2 \\
%  $\rho_2 $ & 0.04 & 0.02 \\
%  $\phi_1$ & 0.3 & 0.2 \\
%  $\phi_2$ & 0.3 & 0.2 \\
%  $\eta$ & 0.02 & 0 \\
%  lr & 0.0001 & 0.0001 \\
% \hline
% \end{tabular}
% \end{center}
% \end{table}

% In both scenarios, we observed that the solution found in the previous training stage preferred an equilibrium where the joints were not so in line, causing useless energy consumption. Therefore, we reinforced the action penalty and increased the bonus for the cosine of the angle between the joints when stabilized.

% This second stage of training, thanks to the tuned reward function with hyperparameter reported in \cref{finetuning_parameters} and the slower training starting from the solution found by the previous training stage, led to agents with performance scores of $0.501$ for the acrobot and $0.511$ for the pendubot.

\subsection{Evolutionary strategy agent}
The Soft Actor-Critic (SAC) framework encompasses three primary neural networks: a policy network and two Q-networks. These networks are parametric models; hence, they can be optimized through zero-order methods such as evolutionary strategies. Such strategies excel in optimizing systems that are either highly ill-conditioned or non-differentiable.

Subtle modifications to the policy parameters can yield considerable disparities in the behavior of controllers, consequently affecting performance metrics drastically. The optimization of SAC is inherently non-stationary, as the policy updates are executed through Stochastic Gradient Descent (SGD) on the evolving Q-functions. However, employing evolutionary strategies directly from the competition score proves challenging due to the lack of immediate feedback unless the model stabilizes the pendulum.

To address these issues, we propose a multi-stage training approach. The first stage is the SAC training reported in the previous sections, and the last stage employs the evolutionary strategy SNES, which begins from the policy refined during SAC's fine-tuning phase. Such policy is finetuned with SNES directly on the competition goal. SNES effectively bridges the gap between the surrogate reward function utilized in training the SAC agent and the original score metric from the competition.

Despite its advantages, this methodology is highly contingent on the starting point of the optimization. The resultant policy tends to remain proximate to that of the initial SAC controller, thereby constraining its potential for variation. Nevertheless, we anticipate that the initial SAC agent has approximated a near-optimal policy, with a discrepancy amenable to further optimization via SNES. Notably, SNES requires minimal hyper-parameter tuning, with the key parameters being the variance $\sigma$ of the population, set at $0.01$, and the population size, established at $40$.

However, the robustness of the agent is a critical consideration, as SNES tends to converge the agent's posterior to a Dirac Delta Distribution, which may exhibit fragility under perturbations. To mitigate this, we perturb the pre-$tanh$-action with noise from $\mathcal N(0, 0.1)$ and then apply the $tanh$ activation. 
% Without this noise, training with SNES for 100 steps resulted in agents achieving scores of $0.55$ for the acrobot and $0.545$ for the pendubot. While these scores surpass all baselines, the models demonstrated considerable vulnerability, with robustness scores approximately $0.2$.

% Conversely, the more robust agents reported in this study attained scores of $0.523$ and $0.516$ for the acrobot and pendubot, respectively.

The best agents found by SNES finetuning achieve a score of $0.524$ for the acrobot, and $0.596$ for the pendubot, while minimally impacting the robustness scores, which decreases from $0.700$ to $0.692$ for the acrobot, and from $0.800$ to $0.796$ for the pendubot.

\begin{figure}[thpb]
      \centering
      \includegraphics[width=0.9\columnwidth]{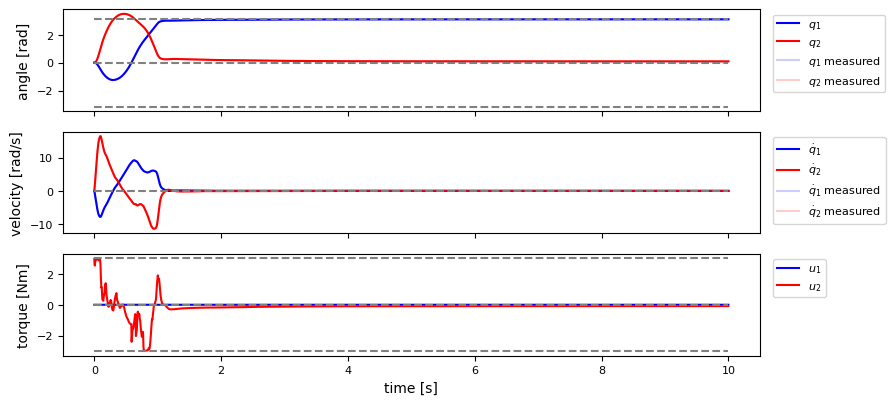}
      \caption{State and input evolution of the final controller for the acrobot task}
      \label{acrobottrajectory}
\end{figure}

\begin{figure}[thpb]
      \centering
      \includegraphics[width=0.9\columnwidth]{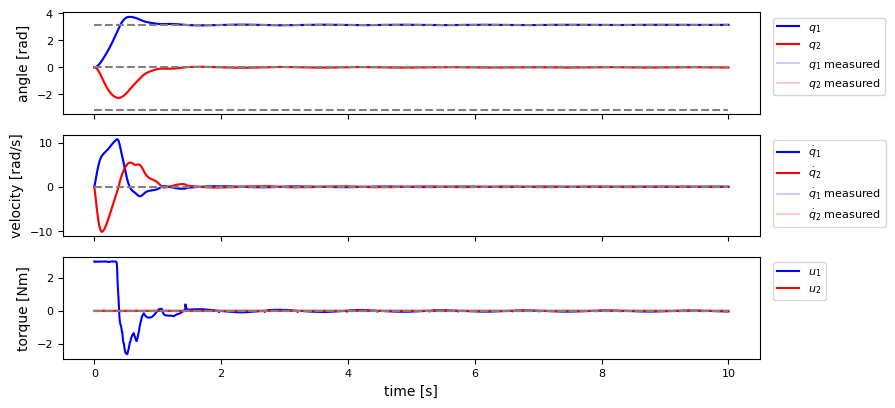}
      \caption{State and input evolution of the final controller for the pendubot task}
      \label{pendubottrajectory}
\end{figure}
\begin{figure}[thpb]
      \centering
      \includegraphics[width=0.49\columnwidth]{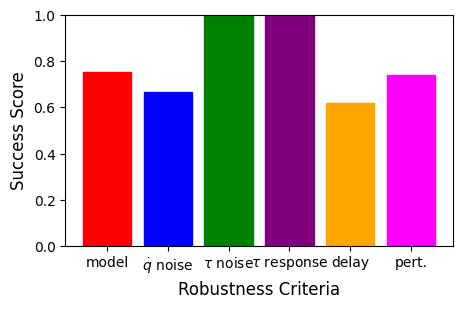}
      \includegraphics[width=0.49\columnwidth]{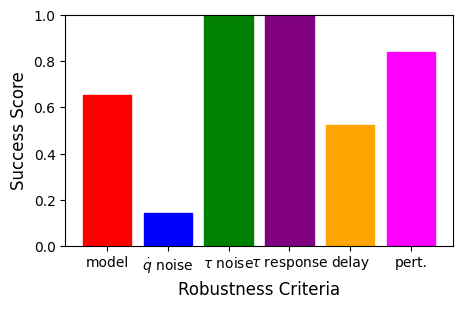}
      \caption{Robustness benchmark results for pendubot (left) and acrobot (right)}
      \label{robustplot}
\end{figure}

\section{Experiments}
For the experiment side of this project, all code was developed in Python. For SAC, we relied on Stable Baselines 3, implemented with PyTorch. For the evolutionary strategy training, we used the EvoTorch library. The code was run on a computer with an AMD Ryzen Threadripper 1920X 12-Core Processor and three Nvidia Titan V GPUs. The evolutionary algorithm was run directly on the CPU.
% For the SAC training, the environment was updated every $10 \, \text{ms}$, whereas in the evaluation and also for the evolutionary strategy, we stuck with the required $2 \, \text{ms}$ update frequency.
During training with SAC, the policy applies control with a frequency of $100 \, \text{Hz}$, whereas in the evaluation and also for the evolutionary strategy, the control frequency is $500 \, \text{Hz}$. 
The learned trajectory can be seen in \cref{pendubottrajectory} for pendubot and in \cref{acrobottrajectory} for acrobot. Regarding the robustness of the agents, the scores for the singular test for pendubot are reported in \cref{robustplot}.
The overall performance results are reported in \cref{tab:robustness_comparison_pendubot} for the pendubot setting and in \cref{tab:robustness_comparison_pendubot} for the acrobot setting.

\begin{table}[h!]
\centering
\begin{tabular}{|c|c|c|c|}
\hline
\textbf{Controller} & \textbf{Robustness} & \textbf{Performance} &\textbf{Avg.}\\
\hline
\textbf{Evolutionary SAC} & \textbf{0.692} & \textbf{0.524} & \textbf{0.608}\\
TVLQR & 0.607 & 0.504 & 0.555\\
iLQR MPC & 0.343 &0.345& 0.343\\
iLQR Riccati Gains & 0.138 &0.396& 0.267\\
\hline
\end{tabular}
\caption{Controller Comparison for the Acrobot}
\label{tab:robustness_comparison_acrobot}
\end{table}

\begin{table}[h!]
\centering
\begin{tabular}{|c | c | c |c|}
\hline
\textbf{Controller} & \textbf{Robustness} & \textbf{Performance} &\textbf{Avg.}\\
\hline
\textbf{Evolutionary SAC} & \textbf{0.796} & \textbf{0.596} & \textbf{0.696}\\
TVLQR & 0.767 & 0.526 & 0.646\\
iLQR MPC & 0.674 & 0.353&  0.513\\
iLQR Riccati Gains & 0.255 & 0.536 & 0.395\\
\hline
\end{tabular}
\caption{Controller Comparison for the Pendubot}
\label{tab:robustness_comparison_pendubot}
\end{table}

\section{Real Robot Stage}
For the real robot testing stage of our study, we deployed the controller initially trained in simulation. However, despite promising results in the virtual environment, the controller did not demonstrate sufficient robustness on the physical system. This outcome was likely due to several discrepancies between the real robot and the simulated model, including variations in masses, lengths, frictional effects, and additional unmodeled phenomena such as sensor noise, torque delays, and inconsistencies in system responsiveness.
\par To bridge this gap, we retrained the policy, systematically introducing stochastic elements into the model to better capture these real-world dynamics. This process included adding uncertainty to physical parameters and simulating environmental noise. In this context, we found that SAC was effective in producing robust controllers capable of handling this increased variability. Although we explored improvements with SNES, the limited interaction time with the physical system constrained SNES's effectiveness.
\begin{figure}[thpb]
      \centering
      \includegraphics[width=0.9\columnwidth]{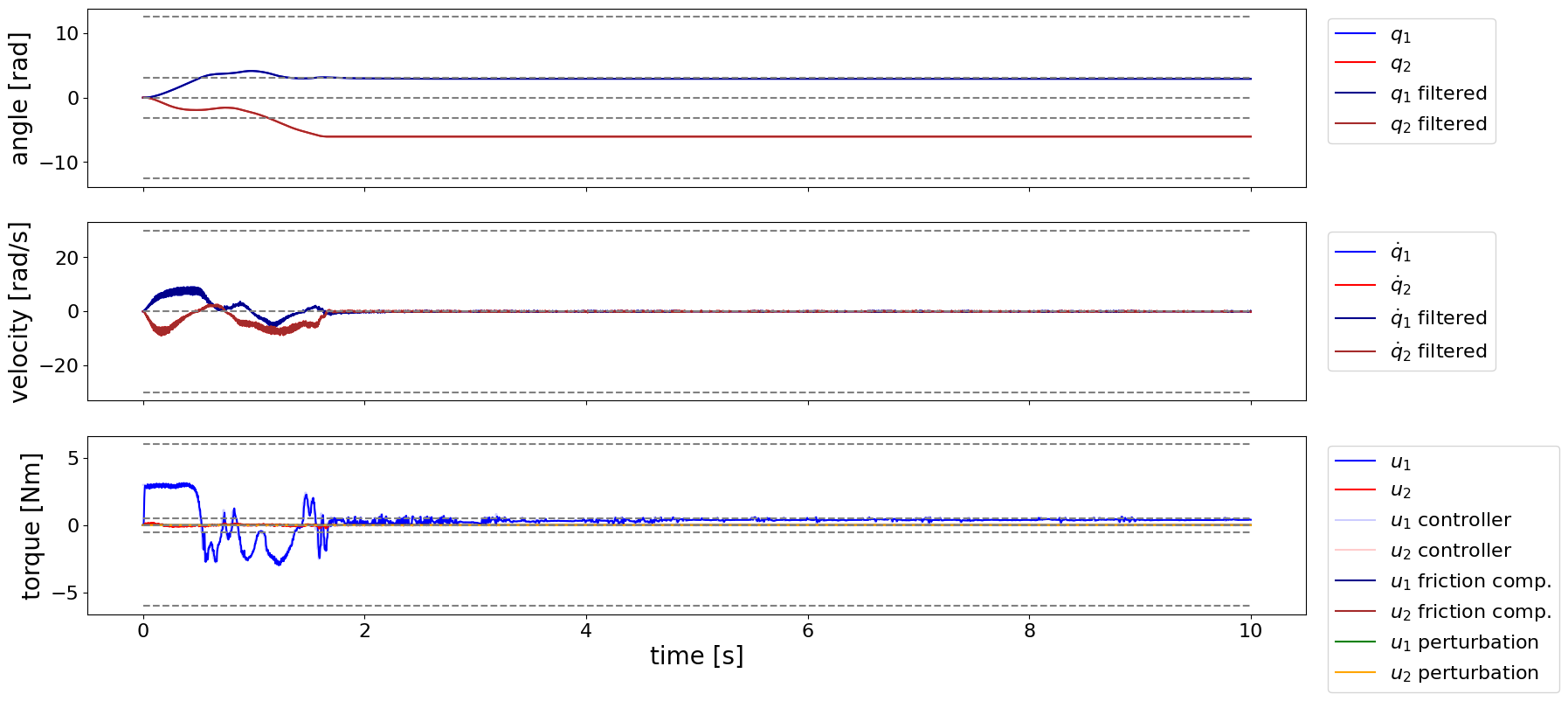}
      \caption{State and input evolution on the unperturbed pendubot system.}
      \label{unperturbed_traj}
\end{figure}

\begin{figure}[thpb]
      \centering
      \includegraphics[width=0.9\columnwidth]{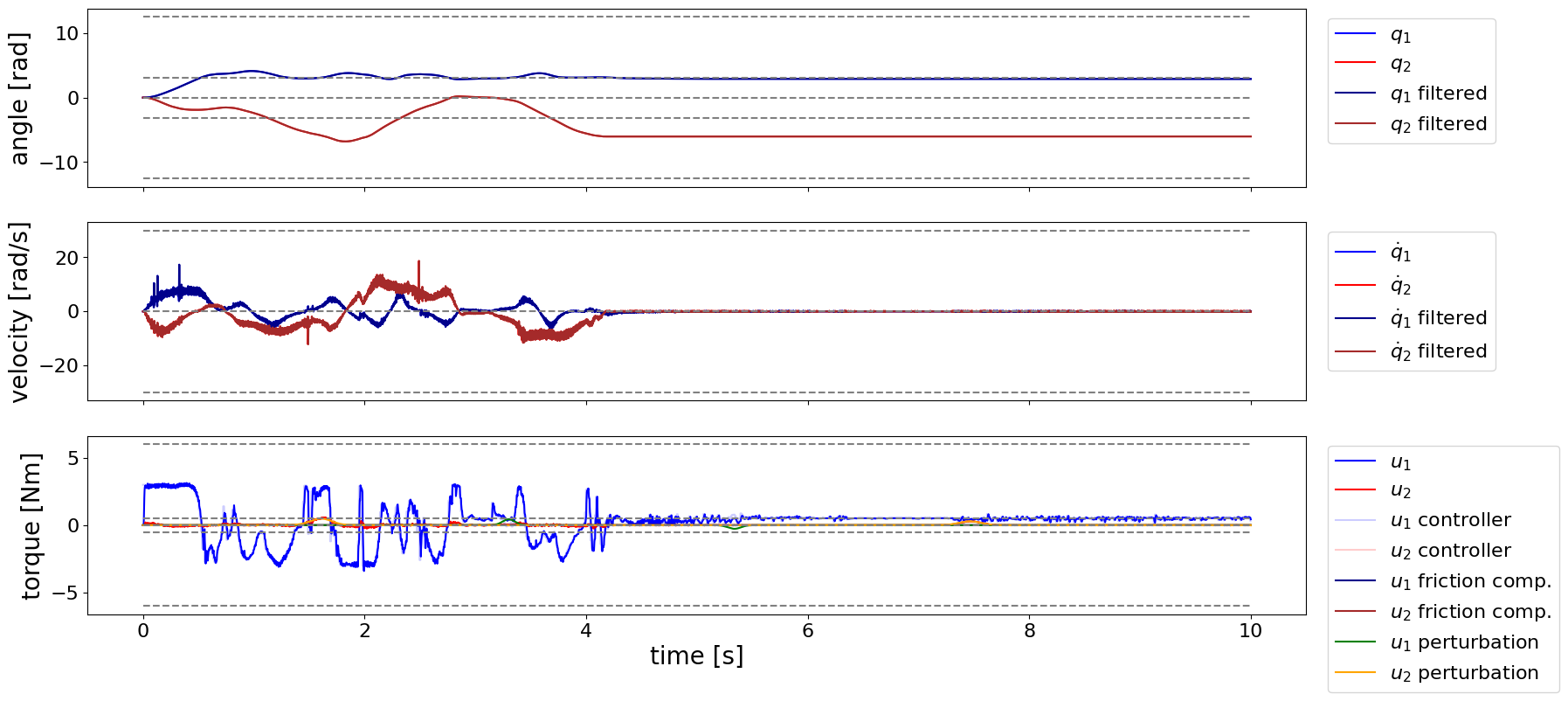}
      \caption{State and input evolution on the perturbed pendubot system.}
      \label{perturbed_traj}
\end{figure}

\begin{figure}[thpb]
      \centering
      \includegraphics[width=.6\columnwidth]{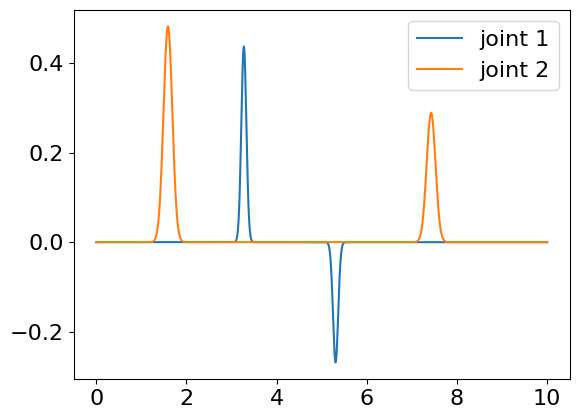}
      \caption{Example perturbations applied on both joints.}
      \label{perturbations}
\end{figure}

We evaluated the resulting controllers in two settings on the Pendubot system. In the first (Figure \ref{unperturbed_traj}), the task involved controlling the robot without external perturbations, yielding an 80\% success rate. In the second ((Figure \ref{perturbed_traj}), we introduced random perturbations on the joints to simulate unexpected disturbances. Here, the success rate varied between 30\% and 80\%, indicating the sensitivity of the controller to perturbations. This phase underscored the challenge of achieving consistent performance across both controlled and unpredictable environments, a critical step in developing reliable real-world robotics controllers. The data can be found at \url{https://github.com/MarcoCali0/RealRobotAIOlympics}

\section{Conclusions and future works}
In this work, we explored a pure end-to-end Deep Reinforcement Learning approach for the AI Olympics challenge. In both settings, our approach surpasses all the baselines. Many other solutions integrated LQR controllers for the stabilization part of the system. Such an introduction allowed them to have faster and more effective training since it addresses the unstable equilibrium point in a more direct approach. Such integration can also be applied to the solution explored for this work. One future line of research is to include an LQR controller for the control near the unstable equilibrium and possibly fine-tune that part with the evolutionary approach. Another route is to redesign the reward function to enable lower torque values to obtain lower energy and efficient controllers.

%%%%%%%%%%%%%%%%%%%%%%%%%%%%%%%%%%%%%%%%%%%%%%%%%%%%%%%%%%%%%%%%%%%%%%%%%%%%%%%%

\end{document}